# FRIDDY MULTIAGENT PRICE STABILIZATION MODEL

Abdelrahman Elsharawy

In a multiagent network model consisting of nodes, each network node has an agent and priced Friddy coins, and the agent can buy or sell Friddy coins in the marketplace. Though every node may not effectively have an equal price during the transaction time, the prices has to reach an equilibrium by iterating buy and sell transaction on a macro level. First, we present a protocol model in which each buying agent in the network (buyer or seller) makes a bid to the lowest priced coins in the part of the network they are at, and that could be an exchange for instance; and each selling agent selects the highest bid, if any. Second, using our model we derive a condition to stabilize price. We also show the equilibrium price can be derived from the total circulating fund and the total Friddy coins for any network. This is a special case of the Fisher's quantity equation[1], and that is the measure we will use to gauge the accuracy of our model. In this paper we will test the best bidding strategy is available to our coin protocol. Third, you will see that we have analyzes stabilization time for what we call path and cycle networks. Finally, simulate market experiments to estimate the tame it should take to stabilize the price of Friddy coin, it's important to note the larger number of market bidders of our coin the more there is an impact on the spreading of funds and the more the model should be tested to understand its ability to stabilize the coin in the ecosystem. This model has shown strong to calibrate the network topologies impact on price stabilization.

## 1. Introduction

**Background.** Conventionally, the topic of currency price stabilization was always researched from a microeconomics point of view [22]. The research relies on supply demand, to as a control of price in the perspective of economics trend. In supply demand curves, if the price is higher than an equilibrium, then simply there is excess that means the price should move to closer to the equilibrium point. In our coin context equilibrium price, the quantity of Friddy coins sought by consumers is equal to the quantity of Friddy coins supplied in the network. And that also means no actors in the network has an incentive to alter price or quantity at the equilibrium. Naturally, in this case the price determined by a market mechanism, simply named the supply-demand theory.

This approach, however, cannot control individual behavior in the marketplace. For instance, suppose that Friddy coins begin at a high market desire point. Under this assumption people would want to buy them immediately even if offered at higher price, and other people want to buy them later at a low price. And since the sum of all those behaviors decide the total demand value, it is virtually difficult to estimate the value by intuition. This model reconsiders the approach in a way the total individual behavior from an overall market mechanism using a multiagent approach, where each market (buyer, seller) agent makes a decision without considering overall advantage/disadvantage. Lets take an example: a game with a minor population and a high base network game that has already attracted attention. The game with a minority base contains non-cooperative, odd number of users who make binary bids (buy or sell) only, in this game hypothesis all binary bids happen in synchronous fashion. The user's bids are made by user deciding best strategy among several ones initially given. Users aim to choose the group of the minority game population because they could potentially make a profit by sharing a great resource. The game continues infinitely and several interesting features, e.g., the cooperation of most agents are known. The model may be considered as a stock market if we incorporate a stock price in the game.

---

[1] The Fisher Equation lies at the heart of the Quantity Theory of Money. MV=PT, where M = Money Supply, V= Velocity of circulation, P= Price Level and T = Transactions. T is difficult to measure so it is often substituted for Y = National Income (Nominal GDP). Therefore MV = PY where Y =national output.

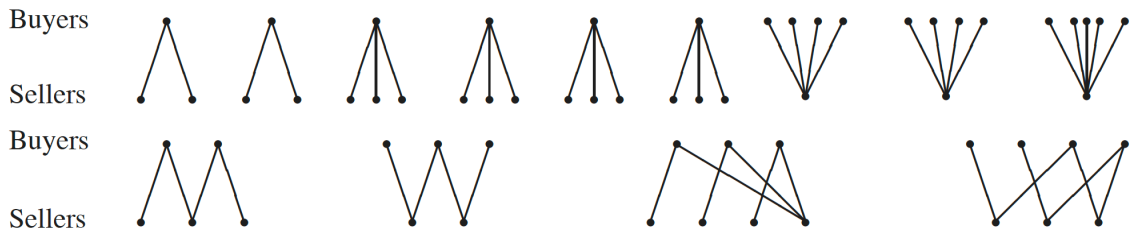

Network formation games [10, 13] contain (non-)cooperative agents who make binary decisions (Buy/Sell), whether to form or stop links. The decisions are made by considering benefits and costs of linking or transacting. Any agent aims to receive the maximum payoff supplied by connected agents and that is the natural instinctive action. The network formation game starts with an empty network, position 0 , forming or severing links by agents on both sides of participating or not, and selling or buying, and ends with a stable state. The model is molded around as a social network formation behavior and what networks are stable in the sense that every agent enjoys maximal satisfaction, and their actions are based on their satisfaction level.

From another dimension, there is the hypothesis of self-stabilization, this concept shares characteristics with the multiagent games in the sense that each network agent or participant's behavior impacts the entire system's outcome or price. The self-stabilization has been originally studied as recovery from transient faults in distributed systems [6]. Human behavior much as the behavior or liquid in physics starts from any initial state, to self-stabilizing algorithms eventually lead to a legitimate state without any aid of external actions.

**Problem.** There are several problems in the previous models. First, it is unclear with what price two agents trade a coin in the marketplace. Some model adopt the middle of maximum buying price and the minimum selling price [21], while other models adopt a log price[3]. However, there is no definite reason on why use one of them. Another problem, that distinct prices reaching an equilibrium is not mathematically modeled for us to measure the probability, speed or mechanics. As far as we know, there is yet to be a multiagent model which works well with the supply-demand theory. Third, it is not clear at what price the distinct prices reach after trading. This is true for both the price produced by two agents in the network and for the equilibrium price produced by all the agents in the network. Fourth, there no previous work that explains the network topologies impact on the stabilization of the market price to the point of equilibrium. Motivated by that I am trying to drive or construct our original multiagent model.

**Solution.** This multiagent network model [15, 16], each agent repeatedly makes auctions and the price for Friddy coin eventually reaches an equilibrium. The network algorithm relation enables us to use an auction and a local trading. The network model consists of nodes and edges as geographic groups and their links to closest neighbors, respectively. Each node contains only one agent who represents people in that particular geographic group. Agents who want to buy Friddy coin make a bid (attempts to transact) to the lowest-priced node, if any, in the neighborhood. Then, agents who want to sell the Friddy coin accept the highest bid. That is, we first consider the local set of agents, and then put them together over the network. This single act delays the recommendation of the farmost prices and priorities the price that is closes to the desired equilibrium price.

With all the stated problems above, we are able address them with answers using few techniques. First, since our model incorporates the principle of an auction, it is clear to determine the trading price. We give a tentative bidding rule in our protocol, and then refer to the best bidding strategy. Second, reliant of the concept of self-stabilization, we will apply some proof techniques. We give the condition of convergence and prove its correctness. Third, since our model assumes the relation among funds, quantity and price at each node, it is clear to derive an equilibrium price after trading. We can confirm that the result coincides with the Fisher's quantity equation. Fourth, since this model contains the network structure, its bart of this model investigate the impacy of network topologies on the price self-stabilization. This model is confirmed to be able derive the stabilization time for path and cycle networks. Finally, in the evaluation of the time stabilization, the number of bidders and the spread of funds for several networks are displayed by simulation.

**Related Work.** You can find more information about the classical theory of price determination in microeconomics is introduced, in [22, 23]. There are globally many economic network models that are published as well, please check [2, 10, 14]. Those models contain a bipartite structure like in [10, 14] or in other case traders play intermediary roles [2]. The model that includes agent-based stabilization has been discussed in [1, 11]. Unlike, in our coin model the ideas is to use mobile agents for the purpose of stabilization. The agents in our model attend sealed bid auctions, that we will manage on chain. For more supported research on the sealed bid auction you'll find arguments in [17, 20]. You will find it useful in designing protocols by what price we should make a bid. There are different kinds of game theoretic methods that appeared in self-stabilization, such as time complexity analysis [7], relationships between Nash equilibria and stabilization [5, 12], and strategies with optimal complexity [9]. In this specific model the method appends an auction approach to such a trend. Our protocol can be concorded a consensus algorithm. The consensus algorithm is presented in [18], and its self-stabilizing version is presented in [6, 8].

**Contributions.** This new network model for economic agents assumes that each agent can buy and sell our coin in the "neighborhood" in our case an example of a "neighborhood" is an exchange or Friddy platform. First, we present in the protocol where each agent always offers a fixed price without considering other bidders' strategies. Then, assume an equilibrium price (1:1 USD) and the availability of the best bidding price. This model also derive the stabilization time during high surge abnormalities up and down in the price. Finally, you will see that we have performed simulations in experiments to understand the effects of network topologies or what we call water surface fluctuations.

Then in section 2 we present our model. Section 3 shows a sufficient condition for stabilization, and then discusses an equilibrium price and the best bidding strategy. Section 4 analyzes the stabilization time for path and cycle networks. Then, Section 5 shows some simulation results. Finally, Section 6 concludes the paper.

## 2. Model

Our system can be represented by a connected network $G = (V, E)$ or Network = (Nodes, Edges), consisting of a set of nodes $V$ and edges $E$, where the nodes represent cities and a pair of neighboring nodes is linked by an edge. Let $N_i$ be a set of neighboring nodes of $i \in V$. We assume that each node $i \in V$ has Friddy coins and their initial price may be definite. The total amount of Friddy coins does not change. Let $p_i(t)^*$ be the price of Friddy coins at node $i$ for the time step $t \in T = [0,1,2,...]$. Each node $i \in V$ has exactly one staying agent $a_i$ who buys/sells Friddy coins in the neighborhood in this case a marketplace or exchange. Each agent $a_i$ has funds (money) $f_i$ and the quantity $q_i$ of Friddy coins. The price $p_i$ is determined by the relation between the quantity of Friddy coins and the buying power, called a supply-demand balance. So, we simply assume two properties at each node. First, the price is proportional to the amount of funds for constant Friddy coins, because it shows the relationship between money supply and inflation. Second, the price is inversely proportional to the amount of Friddy coins for constant funds, because it shows the demand curve. That is,

$$p_i = \frac{f_i}{q_i} \quad (2.1)$$

The buy operation is executed as follows. Each agent $a_i$ assigns a value $v_i^h(t)$ to the Friddy coins of any neighboring node $h \in N_i$, where the value means the maximum amount an agent is willing to pay. Agent $a_i$ compares its own Friddy coins price $p_i$ with any neighboring price in $N_i$. If the cheapest price in $N_i$ is $p_h (< p_i)$, agent $a_i$ wants to buy it and makes a bid $b_i^h (p_h < b_i^h)$ to node $h \in N_i$. For simplicity, we consider $v_i^h(t) = p_i(t)$ for any $h \in N_i$ because he can buy it at price $p_i(t)$ in his node [17].

The sell operation is executed as follows. After accepting bids from $N_h$, agent $a_h$ contracts with $a_i$ in $N_h$, an arbitrary one of agents who made the highest bid $b_i^h$. Then, ah passes his Friddy coins to (receives money from) the contracted agent $a_i$ until the price $p_h(t+1)$ equals to $p_i(t+1)$ derived from the supply-demand balance. In thise case we neglect the carrying cost of Friddy coins rather focus on the change of price of the coin. Notice the price of Friddy coins is updated at each time step. Each node $i \in V$ has a state $\Sigma_i$ represented by a tuple — the Friddy coins and the funds $(q_i(t), f_i(t))$.

---

∗We sometimes denote the price by pi if time t does not matter.
∗We sometimes abuse agent $a_i$ and node $i$ like $a_i \in N_j$.

In this synchronous model, every agent periodically exchanges / attempts to transact and knows the states of neighboring agents or price. We call the state of all nodes a configuration. We describe the set of all configurations as $\Gamma = \Sigma_1 \times \Sigma_2 \times \cdots \times \Sigma_{|V|}$. An atomic step consists of reading the states of neighboring agents, a buy / sell operation, and updating its own state. Then, an atomic step changes a configuration $c_t \in \Gamma$ to $c_{t+1} \in \Gamma$. An execution E is a sequence of configurations $E = c_0, c_1, ..., c_t, c_{t+1}, ...$ such that $c_t \in \Gamma$ changes to $c_{t+1} \in \Gamma$.

In this protocol model, called **Net-Biding**, where each agent $a_i$ makes a bid

$b_i^h (p_h(t) \leq b_i^h \leq p_i(t))$ to agent $a_h \in N_i$ with the lowest price in the neighborhood. Let $c (\geq 1)$ be a constant rate, called a bid parameter, so that the bid $b_i^h$ lies between $p_h(t)$ and $p_i(t)$. For simplicity, we assume the price, or the bid may not be an integer, and we ignore a minute difference between neighboring two prices, such as $|p_i - p_j| < 1$ for $(i,j) \in E$.

**Net-Biding**

Each agent $a_i$ makes a bid

$$b_i^h(t) = p_h(t) + \left(\frac{p_i(t) - p_h(t)}{c}\right) \quad (2.2)$$

A node in the network $h \in N_i$ is the best for an agent to transact with if it has the lowest-priced Friddy coins. An agent contracts with a neighboring node $a_i$ that offers a high bid. The Friddy coins and funds are exchanged as long as they have prices that are lower than those of the other node $i \in N_h b_i^n(t)$. They will continue to exchange until one of them reaches its maximum bid price. If they stop exchanging, then their prices will be determined by equation (2.1) at time t+1.

(priority rule:) If concurrent buy $(b_h^j$ to $j \in N_h)$ and sell $(b_i^h$ from $i \in N_h)$ operations occur at agent `ah, he gives priority to the sell over the buy.

Friddy coins are left. The price is $(p_1(t), p_2(t), p_3(t), p_4(t)) = (50, 110, 70, 10)$. An agent wants to buy the lowest priced Friddy coin at node $h \in N_i$. This means that the price of the Friddy coin at node 4 must be lower than $p_i$. The agent who wants to buy has bid with a price of 110 for node 4

This is about an auction where agents bid to buy something $a_3$ ( i.e., $b_2^1 > b_3^1$). The first agent $a_1$. who bids the most wins and gets the thing being sold. That one pays for it with coins that they bought ahead of time. The first agent who bids the most wins, so they get to pay for it in a way that they choose themselves. Let x be how many Friddy coins that one has to spend on buying before bidding on anything else. If someone has more than x Friddy coins then they can't use them all up before bidding on something else because if you do, then your bid will be too small and you will lose and not get what you wanted!

$$\frac{1000 + 80x}{20 - x} = \frac{2200 - 80x}{20 + x}$$

This gives $x = 3.75$ and hence $q_1 = 20 - x = 16.25$, $q_2 = 20 + x = 23.75$, $f_1 = 1000 + 80x = 1300$, $f_2 = 2200 - 80x = 1900$, and $p_1 = (1000 + 80 \times 3.75)/(20 - 3.75) = 80 = p_2$.

At time At time $t' = t + 1$, the prices become $(p_1(t'), p_2(t'), p_3(t'), p_4(t')) = (80, 80, 70, 10))$.

Since the price of p1 was changed, agent $a_1$ 's bid $b_1^4$ is is made again as $80 + 10)/2 = 45$.. You can make $b_2^3$ and $b_1^4$ at the same time because they are independent.

## 3. Correctness and Properties

In this section, we describe how our protocol works. First, we show that it is correct. Second, we find the price for our protocol assuming that there is an equilibrium price. Third, we show which prices are available to us and how much they cost.

### a. Correctness of our protocol

Some people have a lot of money and some people have very little. This is what makes prices different. Our concern is whether the prices will reach an equilibrium price. To find out, we define legitimacy as follows:

### b. Definition (legitimate configuration).

Every node has a fair price and everyone has updated their prices $t$ to $t + 1$. We call this a legitimate configuration, or in other words, a fair situation. There can't be any deadlocks $C_t \subseteq V$ because this is not an unfair situation where some nodes have more coins than others do.

### c. Lemma

Net-Bidding is deadlock-free. This means that if there are nodes in a configuration, it cannot be an illegitimate configuration $C$. To see this, take any bidding request and show that it doesn't form a cycle with other requests. Then, suppose the configuration is illegitimate at time t. There must be a pair of neighboring nodes $i, j \in V$ with $p_i(t) = \max_{m \in N_j} p_m(t)$ and $p_j(t) = \min_{m \in N_i} p_m(t)$, where $p_i(t) - p_j(t)$.
In this case, their price difference must be greater than the value of the node they are connected to and smaller than the value of their neighbor's neighbor.

If the bids are not in the same order as values, we do not have price stabilization. Proof: Let agent $j \in N_h$ always make a bid with the lowest price in the neighborhood. If he always loses and thus does not change prices, it is because $(b_i^h < b_j^h)$ for all other nodes $v_i^h \geq v_j^h$. we cannot guarantee the price stabilization.

Let $P^{max}(t) = \max_{i \in C_t} p_i(t)$ be the highest price in $C_t$, and $P^{min}(t) = \min_{i \in C_t} p_i(t)$ be
Suppose that people's bids are in the same order as their values $diff(t) = P^{max}(t) - P^{min}(t)$. If any contract price lies between the buyer's price and the seller's price, then price stabilization occurs.

Proof. Let $v_i^h(t) = p_i(t) = P^{max}(t)$. Let node $h$ with $\min_{h \in N_i} p_h(t)$ have the minimum price
In the neighborhood, there are many agents. All bids are ordered in order of value. Agent $a_i$ makes the highest bid in $N_h$. Then agent $a_i$ can contract with $a_h$ to buy. It means $P^{max}(t) = p_i(t) > p_i(t + 1)$. Assuming there no other agents make bids greater than $P^{max}(t)$, we have $P^{max}(t) > P^{max}(t + 1)$. The similar argument holds for $P^{min}(t)$. we have

$$\text{diff}(t) > \text{diff}(t + 1). \quad (3.1)$$ This means price stabilization occurs.

If you do not have a contract between the price of the home and the buyer, then you cannot guarantee that there will be price stabilization. This is because the inequality does not hold. As long as $a$ makes a bid of $v_i^h(t) = p_i(t)$ for any neighboring node $h \in N_i$, then Net-Biding satisfies this condition.

### d. Equilibrium price

The price of cryptocurrencies in a network will always be determined by the total amount of funds and Friddy coins in that network. If there are not enough, it can result in an outflow from G-value into another ecosystem which has higher cryptocurrency prices, where demand exceeds supply.

$$P^e = \frac{F}{Q}$$

I know that a network topology is the pattern of connections between nodes in a graph. I also know that the price of Friddy coins at node $i$ is $p_i = f_i/q_i$.. If prices are different at each stabilization, then they will not always be equal.

That is, $P^e(t) = p_i(t) \neq p_i(t') = P^e(t')$ for time $t$ and $t' (t \neq t')$ holds. Since $f_i = p_i(t)q_i$ and $f_i' = p_i(t')q_i'$ hold for any node $i$, where $F = \sum_i f_i = \sum_i f_i'$ we have

$$p_i(t) \cdot \sum_i q_i = p_i(t') \cdot \sum_i q_i'$$

Since the total amount of Friddy coins Q is the same, we have

$$Q = \sum_i q_i = \sum_i q_i'$$

Thus, we have $p_i(t) = p_i(t')$, a contradiction. Therefore, the equilibrium price $P^e$ is identical for each stabilization.
Next, since $f_i = P^e \cdot q_i$ holds for every node $i$, the total funds sum up to

$$F = P^e \cdot Q$$

Thus, we have $P^e = F/Q$.
In Fisher's quantity equation [19] $FV = P^e Q$ if the velocity of funds V equals to 1. This means the correctness of our assumption at each node.
Let $p^e(t+1)$ be the temporary, shared price of nodes $i$ and $j$ reached by trading exhaustively for a contract between $t$ and $t+1$. From Theorem 3.2, we have the following corollary.
**Corollary** Let $q_i(t)$ and $q_j(t)$ be the quantity of Friddy coins, and $f_i(t)$ and $f_j(t)$ the funds before the trade, respectively. After the trade, the shared price will be

$$p^e(t+1) = \frac{f_i(t) + f_j(t)}{q_i(t) + q_j(t)}$$

**Best bidding for constant bidders**

In this part of the paper, we will consider whether Bayesian Nash equilibrium is applicable in our model. It says that if all bids are equal in size then it will be the best strategy. Consider a game where all agents have values between \alpha and \beta and they are independent of each other. If so, then the expected outcome will be $\mathcal{N}$(\alpha+β)\), It means that every person gets what they want, but they may not get it for as long of a time. On average people get about 70% of their bid requests fulfilled with some variation depending on how uniform you think these distributions are--the more random your distribution across the day, then the more likely it is that you will be able to fulfill most of your bids. (α + β ), the better chance everyone not only has their desired amount but also may even surpasses. Then, the distribution function is $F(x) = r(x - \alpha)$, where $r = 1/(\beta - \alpha)$. Let $Y$ be the highest of $B - 1$ values. Then, $Y$ is the highest order statistic of the values. Thus, the distribution function of $Y$, denoted by $G(x)$, is $G(x) = F(x)^{B-1}$. In addition, let $g(x)$ be the density function of $G(x)$. Let $B^h(t)$ be the number of bidders to node h at time t, or simply denoted by B. Agent $a_i$'s best strategy $S(v_i^h)$ against $B - 1$ bidders is known [16, 17] as

$$S(v) = \frac{1}{G(v)} \int_\alpha^v y g(y) dy$$
$$= \frac{1}{\{r(v-\alpha)\}^{B-1}} \int_\alpha^v y r^{B-1}(B-1)(y-\alpha)^{B-2} dy$$
$$= v - \frac{v - \alpha}{B}$$

We now need to see if the strategy when you bid and value is correct. First, suppose Suppose that $v_i^h \leq v_j^h$ holds.

$$b_j^h - b_i^h = S(v_j^h) - S(v_i^h)$$
$$= \left(v_j^h - \frac{v_j^h - \alpha}{B}\right) - \left(v_i^h - \frac{v_i^h - \alpha}{B}\right)$$
$$= (v_j^h - v_i^h)\left(1 - \frac{1}{B}\right) \geq 0.$$

Thus $b_i^h \leq b_j^h$ holds.
Next, for the bidding price,

$$S(v_i^h) - \alpha = v_i^h - \frac{v_i^h - \alpha}{B} - \alpha = (v_i^h - \alpha)\left(1 - \frac{1}{B}\right) > 0.$$

On the other hand, it is clear that

$$v_i^h - S(v_i^h) = v_i^h - \left(v_i^h - \frac{v_i^h - \alpha}{B}\right) = \frac{v_i^h - \alpha}{B} > 0.$$

Thus, we have $\alpha < S(v_i^h) < v_i^h$ In our protocol, Net-Bidding, if each agent follows the rules, then we will obtain a theorem. Theorem 3.3 is what we get when all of the agents follow the rules in this protocol. $a_i$ confronting $B - 1$ bidders repeatedly make a bid to the lowest-price node $h \in N_i$ by strategy

$$S(v_i^h(t)) = v_i^h(t) - \frac{v_i^h(t) - p_h(t)}{B} = p_i(t) - \frac{p_i(t) - p_h(t)}{B}.$$

The prices in our model change over time. It is necessary to know the precise number of bidders, even though it can be assumed that there is a max number of people who have placed bids on each node.

## 4. Analyses

In this part, we will investigate how long it takes to stabilize. First, we focus on the path. Next is the cycle.

### a. Stabilization time for a path

In the sequel, we analyze the stabilization time for a path $(1, \ldots, n)$.. the price movement in a path. Suppose that there is a path $(1,2,3,4,5,6)$ such that $(p_1(0), \ldots, p_6(0)) = (50,57,55,60,56,70)$, where the prices are represented by the nodes position in the vertical direction ). After time 1, several pairs of prices are created and similar movements keep happening. The position of nodes does not change after every other time.
We only look at prices in an ascending or descending sequence so it will take the time to stabilize. We assume $n = 2m$ is even. The following theorem is about how long it takes for a path to stabilize.

**Theorem 4.1.** If network $G$ is a path, the stabilization time of our Net-Biding is $2\tau$ steps, where $\tau$ satisfies

$$\left(\frac{3}{4}\right)^\tau \left(\frac{1}{3}\right)^{n/2+1} \left(\frac{\tau}{n/2+1}\right)^{n/2+1} = 1.$$

Proof. We call a price difference between two nodes a gap. We call the gaps as 1st gap, 2nd gap.. . . in the ascending order of the nodes. Let $d_i(t)$ be the difference of the $i$-th gap $(1 \leq i \leq m)$ at time $t$, where $t$ means every other time here. Then, we have the following recurrences.

$$d_i(t+1) = \frac{1}{4}d_{i-1}(t) + \frac{1}{2}d_i(t) + \frac{1}{4}d_{i+1}(t). \quad (4.1)$$

$$d_1(t+1) = \frac{1}{2}d_1(t) + \frac{1}{4}d_2(t) \quad (4.2)$$

$$d_m(t+1) = \frac{1}{2}d_m(t) + \frac{1}{4}d_{m-1}(t). \quad (4.3)$$

Let $S_{m-j}(t) = \sum_{i=j+1}^{m-j} d_i(t)$ and $S_m(0) = D$. Summing (4.1) from $i = 1$ to m by using (4.2) and (4.3) gives

$$S_m(t+1) = S_m(t) - \frac{1}{4}(d_1(t) + d_m(t)).$$

Since $d_1(t) + d_m(t) = S_m(t) - S_{m-1}(t)$, we have

$$S_m(t+1) = \frac{3}{4}S_m(t) + \frac{1}{4}S_{m-1}(t)$$

If we define the generating function $G_t(z) = \sum_{m \geq 0} S_m(t)z^m$, Equation (4.4) can be written as

$$\sum_{m\geq 0} S_m(t+1)z^m = \sum_{m\geq 0} \frac{3}{4}S_m(t)z^m + \sum_{m\geq 1} \frac{z}{4}S_{m-1}(t)z^{m-1}$$

that is,

$$G_{t+1}(z) = \frac{3+z}{4}G_t(z) = \cdots = \left(\frac{3+z}{4}\right)^{t+1} G_0(z)$$

Since $G_0(z) = \sum_{m\geq 0} S_m(0)z^m = D\sum_{m\geq 0} z^m = \frac{D}{1-z}$ we have

$$G_t(z) = \left(\frac{3+z}{4}\right)^t \cdot \frac{D}{1-z} = \frac{D(3/4)^t(1+z/3)^t}{1-z}$$

Hence, the coefficient of $z^m$ is

$$D(3/4)^t \sum_{k=0}^{m} (1/3)^k \binom{t}{k}$$

$$= D(3/4)^t \left\{ (1+1/3)^t - (1/3)^{m+1}\binom{t}{m+1} - O(1/3^{m+2}) \right\}$$

$$\leq D\left\{ 1 - \left(\frac{3}{4}\right)^t \left(\frac{1}{3}\right)^{n/2+1} \left(\frac{t}{n/2+1}\right)^{n/2+1} \right\}$$

because $m = n/2$. Hence, $S_m(t) = 0$ gives $\left(\frac{3}{4}\right)^t \left(\frac{1}{3}\right)^{n/2+1} \left(\frac{t}{n/2+1}\right)^{n/2+1} = 1$. Since it takes $2t$ steps until convergence, the lemma follows.

## 5. Stabilization time for a cycle

Since a path $(1,2,\ldots,n)$ becomes a cycle if both ends, 1 and $n$, are connected, we can

Since a path $(1,2,\ldots,n)$ We want to connect the two ends with the lowest prices at node 1 and node n. If we do that, then all of the nodes in-between will be one price higher than the previous one. That way, we can move from those two points without changing them. The prices for different items will change depending on the price of other items. If an item's price is low, then it will go up. If something is high, then it will go down. Prices can also change depending on how many units were sold for that item so far. The prices are not static and they keep changing throughout history based on what people buy

We want to know about the center part which controls the time it takes for a cycle. Is it when the lowest price node shifts over to a neighboring intermediate node? We call this difference between prices on either side of that point. (resp. right- neighboring) node and the lowest price node a left-difference (resp. right-difference), denoted by $d_{i-1}$ (resp. $d_i$) . The lowest price If the price is higher on left than right, it will move to the next node. If it is higher on the right than left, it will stay where it is. The lowest priced nodes move until they are at the same level. This always happens eventually.

$$\frac{d_{i-1}}{2} - \frac{d_i}{2} = \frac{d_{i+1}}{2} + \frac{d_i}{2}$$

$$d_i = \frac{1}{2}(d_{i-1} - d_{i+1})$$

If we assume $d_i \approx d_{i+1}$, we have $d_i = d_{i-1}/3$. This means the left-hand side difference is about 3 times as large as the center part difference. Thus, the center part length is at most $3n/4$. So we have the following corollary.

Corollary 4.1. The stabilization time for an n-node cycle is less than that for a $a\frac{3n}{4}$ −node path.

## 6. Simulation

We examined the number of steps for stabilization in Section 5.1. In this section, we proposed two methods of estimating the number of bidders and evaluated them for several types of topologies in Section 5.2. Then, we focused on the spread of funds throughout networks and investigated them for several types of topologies in Section 5.3

In the following experiments, we will describe common constants and parameters. We repeat the experiment up to 500 trials, where a trial ends with equilibrium. We will set the number of nodes from 50 to 500 when others are fixed. at $c = 2$ in (2.2).

Table 1: Common constants and parameters

| Meaning | Values |
| --- | --- |
| Number of trials | 500 |
| Number of nodes | 50 — 500 |
| Amount of funds | 10,000 |
| Amount of Friddy coins | [50, 100] |

## 7. Stabilization time

We study how different topologies with graphs affect price stabilization. We observe four kinds of graphs: a path, a grid, a random k-link graph, and the complete graph. The random k-link graph is defined as a cycle with randomly selected edges. We compare the number of steps until convergence on these graphs.

The time it takes to stabilize is different in graphs with 0-links, 1-links, and 2-links. A path has no extra edges so it will take a long time to stabilize. A cycle (0-link) also has no extra edges and will take a long time. The stabilization times for two 1-links are the same length as the 3 networks because they each have an extra edge.

This is why Corollary 4.1 says that paths and cycles (0 links) limit the spread of prices has many 4-degree nodes. That means a lot of agents are trading at the same time. The complete graph has only one pair of agents, so it takes longer to stabilize than the grid graph does because there are less trade partners.

The random $|V|/10-$link The graph has an intermediate feature between the complete graph and the grid graph. This means that since there are randomly selected edges, it is somewhat unstable.

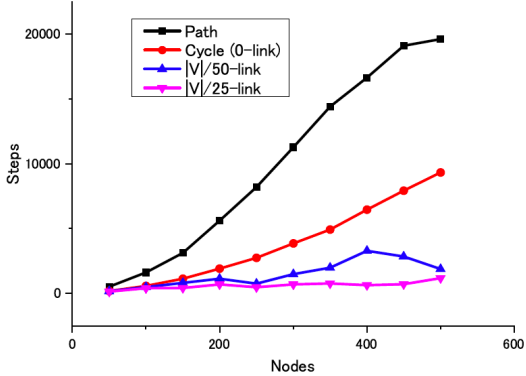

Figure 5: Stabilization time (path vs. k-link)

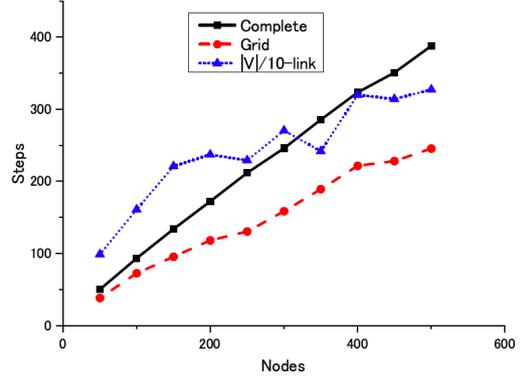

Figure 6: Stabilization time (several networks)

a. Number of bidders
As you read in Section 3.3, it is necessary to know the number of bidders B for the decision of the best bidding price. Net-Biding does not contain such a method, but we give it in the following. We compare two methods which enable us to estimate B. The first is suitable for estimating B and the second is not suitable for estimating B.

b. A method is to use $B$ in the previous step by assuming the number of bidders is available.

c. ne way to estimate B is by assuming that the agent values are uniformly distributed over the same interval. Method 1 uses the previous information and assumes that nothing would suddenly change in time. For example, if agent a_i wants to estimate B, they could use Method 1 because it assumes that each agent value is uniformly distributed over the same interval.

$B(t)$ at node $h \in N_i$. Then $a_i$ just substitutes $B(t-1)$ for $B(t)$. Method 2 uses information about the prices in the area that is close to where your house is. More precisely, let $gap_h$ be the difference between the maximum price in $N_h \cup h$ and the minimum price in $N_h \cup h$. Let $g_h$ be the difference between the maximum price in $N_h \cup h$ and the price of node $h$. Then, agent $a_i$ estimates the number of bidders to node $h \in N_i$ as

$$e_h(B) = \text{int}\left(\frac{g_h}{max(gap_h, 1)} \cdot |N_h|\right) \quad (5.1)$$

where "int" rounds off to an integer and $max(gap_h, 1)$ avoids zero for a denominator.
We have two types of graphs. One is called a sparse graph and the other is called a dense graph. A random person selected which kind of graph to do first. $k-$ link graph with different $k$ values. To evaluate the methods, we use the following expression

$$D = \sum_{i \in V} \left( \frac{e_i(B) - a_i(B)}{|N_i|} \right)^2$$

where $a_i(B)$ is the actual number of $(B)$ We collect the value of $D$ in each step and take an average of them. Notice that D represents a deviation from the actual number of bidders, and its range is $0 \leq D \leq |V|$.
This is the difference between two graphs.. Method 2 is a little
more effective than Method 1 in dense $(k = |V|/1.2)$ networks. On the other hand, Method 1 is much more effective than Method 2 in sparse $(k = |V|/10)$ networks. Since there are many choices in the network, it would be useful to use a method like Method 2.

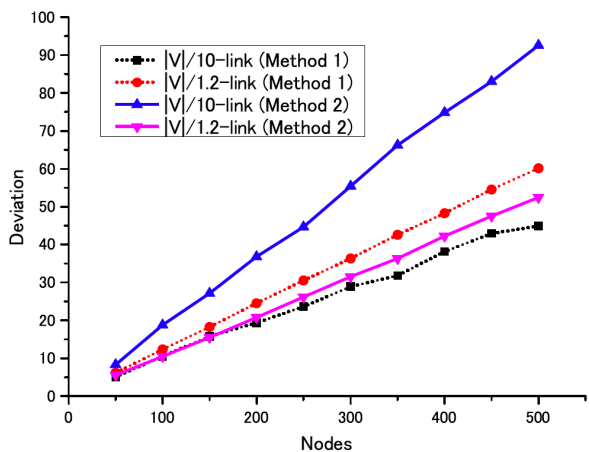

Estimation of number of bidders

The injection of funds in the network is useless.

There are not many buyers and sellers in sparse networks and also not enough money for each person to buy Friddy coins.
Therefore, injecting money is a useless policy.
1. The topology of a network can influence the spread of funds. Sometimes, for example, the number of links in a network can affect how far the funds travel.
2. The number of nodes in a network also affects how quickly funds travel through it.
3. Injections points also affect how fast funds are distributed through a system because they represent opportunities to enter and exit the network as well as access different areas where money is likely stored or being transferred from one point to another which may be more or less secure than others depending on their physical location and whether they are government controlled facilities such as banks, border crossings or airports.
4. The amount of money transmitted into an injection point will determine how many other points 100 for a non-injection point and 30,000 for an injection point. The number of coins depends on where you are in the table. Some numbers are more common than others.

Friddy coins is determined by Equation (2.1).

Table 2: Basic constants

| Meaning | Value |
|---|---|
| Amount of funds | 100 |
| Injection funds | 30,000 |

Table 3: Parameters

| Meaning | Value | Standard Value |
|---|---|---|
| Number of nodes | 50—500 | 300 |
| Number of injection points | 1—10 | 1 |

We know about four types of networks, a path, a grid, and a complete graph. A grid is the type of network where each node has four or more edges connected to it. The complete graph is when every node has an edge connecting them to all other nodes in the network. We use k = |V|/10 in the following experiments.

Max / min funds vs. number of nodes for four networks
Max / min funds vs. number of injection points for four networks

The most ineffective network is the path. It has a node that is very far from where the injection point is. Mostly, it can be used to spread funds. The complete graph and grid have this property because they have an increase of nodes and extra edges, while the path does not. This is because there are many pairs of trading nodes in the grid because they consist of local maximal or minimal price nodes. The complete graph has only one pair of trading nodes with maximum and minimum prices. The grid can do a better job than the complete graph for spreading out funds because it does not have that one pair of trading nodes. We want to change the number of nodes in a network from 1 to 10. We will do this randomly. We will also fix the number of nodes at 300. This graph shows that our new grid with 1/10-links is better than other networks when it comes to spreading funds. This is true for networks with many links. If every node in the network were injected with funds, then the fund-spreading policy would be complete. We can see this more clearly when we look at a network of 300 nodes and change the number of links from 0 to 1500. The graph shows that as the number of injections increases, the point goes quickly to 0. Then it goes more slowly as you add more injections.

## 8. Conclusion

This paper is about a new way to stabilize prices. First, we looked at the system model, which suggests that the price should be proportional to funds and inversely proportional to Friddy coins at each node. We also looked at how this could work with a protocol that moves funds / Friddy coins. We looked at how prices are determined in the market. We found that they depend on how many funds people have and how many coins they have. We also found out that the equilibrium price is when these two values are equal. And we examined a bidding strategy for constant bidders who can use our protocol to bid on properties. Finally, we looked at special cases and their stabilization time

- In summary, our network model shows the following facts.
- The price stabilizes if bids are the same order as values and any contract price is in between buyer's price and seller's price. The equilibrium can be estimated if the price is proportional to funds (inversely proportional to Friddy coins) at each node.
- Denser networks are easy to reach equilibrium because there are multiple paths for spreading the prices.
- For the best bidding, the Bayesian-Nash solution needs the number of bidders B. To estimate B, our Method 2 is a little useful for a dense graph, while our Method 1 is useful for a sparse graph.
- We prefer networks with sparse neighborhoods like the grid instead of networks with dense neighborhoods, such as a complete graph.
- Our goal is to create a good model so that we can study and understand different economic phenomena. We will investigate an asynchronous model and work on developing other protocols.

Abdelrahman Elsharawy
ar@friddy.com
http://www.friddy.com